\documentclass[10pt,twocolumn,letterpaper]{article}
\usepackage{iccv}
\usepackage{times}
\usepackage{epsfig}
\usepackage{graphicx}
\usepackage{amsmath}
\usepackage{amssymb}
\usepackage{multirow}
\usepackage{gensymb}
\usepackage{cite}
\usepackage[breaklinks=true,bookmarks=false]{hyperref}
\hypersetup{
colorlinks=true
}

\iccvfinalcopy 
\def\iccvPaperID{****} 

\newcommand{\setParDis}{\setlength {\parskip} {0.25cm} }
\newcommand{\setParDef}{\setlength {\parskip} {0cm} }

\def\authorBlock{
    Yang Fu\textsuperscript{1}\thanks{Equal contribution}~,
    Shibei Meng\textsuperscript{1}\footnotemark[1]~,
    Saihui Hou\textsuperscript{1,2}\thanks{Corresponding Author}~,
    Xuecai Hu\textsuperscript{1},
    Yongzhen Huang\textsuperscript{1,2}\footnotemark[2]  \\
    \textsuperscript{1~}School of Artificial Intelligence, Beijing Normal University~
    \textsuperscript{2~}WATRIX.AI \\
    {
    \tt\small \{yangfu, mengshibei\}@mail.bnu.edu.cn,
      \{housaihui, huxc1208, huangyongzhen\}@bnu.edu.cn
     }
}

\begin{document}

\title{GPGait: Generalized Pose-based Gait Recognition}
\author{\authorBlock}
\maketitle

\begin{abstract}
    Recent works on pose-based gait recognition have demonstrated the potential of using such simple information to achieve results comparable to silhouette-based methods. However, the generalization ability of pose-based methods on different datasets is undesirably inferior to that of silhouette-based ones, which has received little attention but hinders the application of these methods in real-world scenarios. To improve the generalization ability of pose-based methods across datasets, we propose a \textbf{G}eneralized \textbf{P}ose-based \textbf{Gait} recognition (\textbf{GPGait}) framework. First, a Human-Oriented Transformation (HOT) and a series of Human-Oriented Descriptors (HOD) are proposed to obtain a unified pose representation with discriminative multi-features. Then, given the slight variations in the unified representation after HOT and HOD, it becomes crucial for the network to extract local-global relationships between the keypoints. To this end, a Part-Aware Graph Convolutional Network (PAGCN)  is proposed to enable efficient graph partition and local-global spatial feature extraction. Experiments on four public gait recognition datasets, CASIA-B, OUMVLP-Pose, Gait3D and GREW, show that our model demonstrates better and more stable cross-domain capabilities compared to existing skeleton-based methods, achieving comparable recognition results to silhouette-based ones. Code is available at \url{https://github.com/BNU-IVC/FastPoseGait}.
\end{abstract}

\section{Introduction}
Gait recognition is an essential task in the human identification field.
Compared with existing biometric identification methods, such as face, fingerprint, and iris recognition, it can capture long-distance gait features without the cooperation of subjects.
Existing studies of gait recognition can mainly be divided into two streams, appearance-based~\cite{liang2022gaitedge,hou2022gait,hou2020gait,chao2019gaitset,fan2020gaitpart,fan2022learning,zhang2019comprehensive} and model-based methods~\cite{liao2020model,teepe2021gaitgraph,teepe2022towards,pinyoanuntapong2022gaitmixer,liu2022symmetry,zhang2022spatial}.
Specifically, the appearance-based methods try to directly learn gait features from the silhouette sequences and have been the dominant approach for a long time.
And the model-based methods try to explicitly estimate human body structures (\eg, keypoints or 3D mesh) for gait recognition.
Despite that the performance is inferior to the appearance-based methods at the moment, the model-based methods have their own advantage of being robust to carrying and clothing, which is appealing to practical applications and thus deserves continuous attention.
\begin{figure}[tp]
    \centering
    \includegraphics[width=0.85\linewidth]{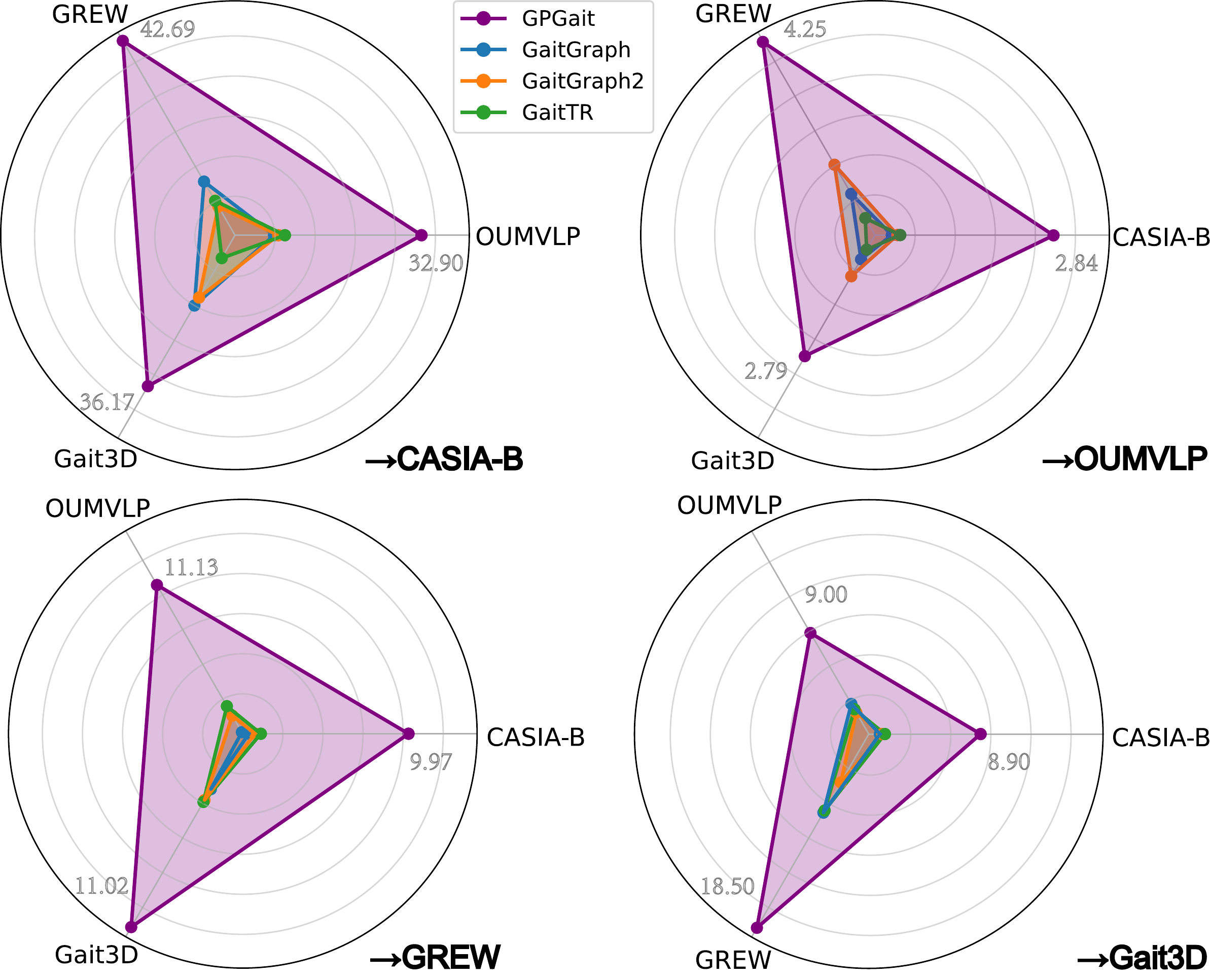}
    \caption{GPGait achieves state-of-the-art generalization performance across four popular gait datasets. In each subgraph, three vertices correspond to the source domain.  Arrows (\(\xrightarrow{}\)) point to the target domain.  Triangles in different colors represent the generalization ability of different models under cross-domain settings.
    }
    \label{fig:radar}
\end{figure}

Model-based methods mostly take human poses (\ie, keypoints) as the input which encodes visual clues for body structure and proportion in an explicit way.
Benefiting from the rapid development in pose estimation~\cite{8765346,alphapose,li2022simcc,8953615} and graph-based models (GCN~\cite{yan2018spatial,song2020stronger} and Transformer~\cite{plizzari2021spatial}), pose-based methods have achieved fairly surprising results in some cases~\cite{teepe2021gaitgraph,teepe2022towards,zhang2022spatial}, \eg, GaitTR~\cite{zhang2022spatial} introduces Spatial Transformer to establish overall spatial relationships between keypoints. The model achieves much higher accuracy than previous pose-based methods~\cite{teepe2021gaitgraph,teepe2022towards} on CASIA-B~\cite{yu2006framework}, even surpassing the accuracy of appearance-based methods~\cite{chao2019gaitset,hou2020gait,fan2020gaitpart,lin2021gait} in clothes-changing conditions.

However, a vital problem is ignored in these research, \ie, generalization ability.
Through a preliminary study as shown in Fig.\ref{fig:radar}, we find that the performance of these methods tends to drastically degrade when testing gait sequences from unseen environments, limiting the application in realistic scenarios.
Our analysis suggests that the decline in cross-domain performance can be linked to various factors:
(1) scale variations due to the distance to cameras, (2) tilt and horizontal views due to the deployment of cameras, (3) offsets within the camera coordinate system.
All these factors can cause intense changes in data distributions, resulting in a dramatic decline in performance when there are variations in cameras and environments.

To promote the research on model-based gait recognition, we aim to design a framework for \textbf{G}eneralized \textbf{P}ose-based \textbf{Gait} Recognition (GPGait) that can effectively improve the generalization ability of pose-based methods.
Particularly, we try to solve the problem from two perspectives: \emph{a human-oriented input that is comparable across different cameras}, and \emph{a part-aware model that extracts fine-grained body features for recognition}.

In terms of input format, we propose a Human-Oriented Transformation (HOT) and a series of Human-Oriented Descriptors (HOD) to obtain a unified and enriched representation.
Specifically, HOT consists of three steps, namely affine transform, body rescale, and body alignment, through which the original skeleton sequences captured in the camera coordinate system are transformed into unified representations in the human-oriented coordinate system.
Then, to enrich the input, we carefully design a module named Human-Oriented Descriptors (HOD) to generate individual-invariant features of bone and angle to explicitly reflect the body proportion and structure.

Regarding the modeling, we argue that the fine-grained learning for different human parts is the key to extracting discriminative gait features and improving the generalization ability.
Although we can obtain a unified representation with HOT and HOD, the uniform gait expression exhibits less variation over time compared to the original pose sequence.
Therefore, it is imperative to capture the local-global relationships between the keypoints, where local features can capture the slight changes in pose and the global ones can represent the entire human structure.
Inspired by the recent progress in the field of gait recognition~\cite{chao2019gaitset,fan2020gaitpart,hou2020gait,chai2022lagrange,lin2021gait} and domain generalization~\cite{fan2022learning}, we design a Part-Aware Graph Convolutional Network (PAGCN) which  can efficiently implement graph partition and local-global relationship construction through mask operations on the adjacency matrix.

To summarize, we make the following three major contributions:
{
\begin{itemize}
    \item In the HOT module, a series of human-oriented operations are proposed to facilitate a uniform input that overcomes problems caused by various environmental covariances. The input is further enriched in the HOD module to explicitly reflect the skeleton structure and movement.
    \item We present PAGCN to achieve efficient graph partition and local-global feature relation extraction under the unified pose representation. With different part-specific masks and a well-designed network structure, the method can not only capture fine-grained features and distinct local relations but also reduce the amount of calculations and the number of parameters required.
    \item Extensive experiments demonstrate that the proposed GPGait framework achieves state-of-the-art generalization results in all scenarios (indoor and outdoor) under cross-domain settings.
          Especially, the result of the cross-domain test on GREW$\xrightarrow{}$CASIA-B outperforms previous methods by a large margin of 34.69\%.
\end{itemize}
}

\section{Related Work}
\subsection{Gait Recognition}
Gait recognition methods can be classified into two main categories: appearance-based~\cite{han2005individual,chao2019gaitset,fan2020gaitpart,hou2020gait,hou2022gait,liang2022gaitedge} and model-based methods~\cite{teepe2021gaitgraph,teepe2022towards,li2020end,li2022multi,xu2023occlusion,liao2020model,liao2017pose} depending on the type of data input. Appearance-based methods usually rely on silhouette sequences~\cite{chao2019gaitset,fan2020gaitpart,hou2020gait,hou2022gait,liang2022gaitedge} or their transformation, such as Gait Energy Image (GEI)~\cite{han2005individual} and Chrono-Gait Image (CGI)~\cite{wang2011human}. Model-based methods, on the other hand, represent the human body as mesh~\cite{li2020end,li2022multi,xu2023occlusion} or a set of  keypoints~\cite{liao2020model,liao2017pose,teepe2021gaitgraph,teepe2022towards}.

\begin{figure*}
    \centering
    \includegraphics[width=\textwidth]{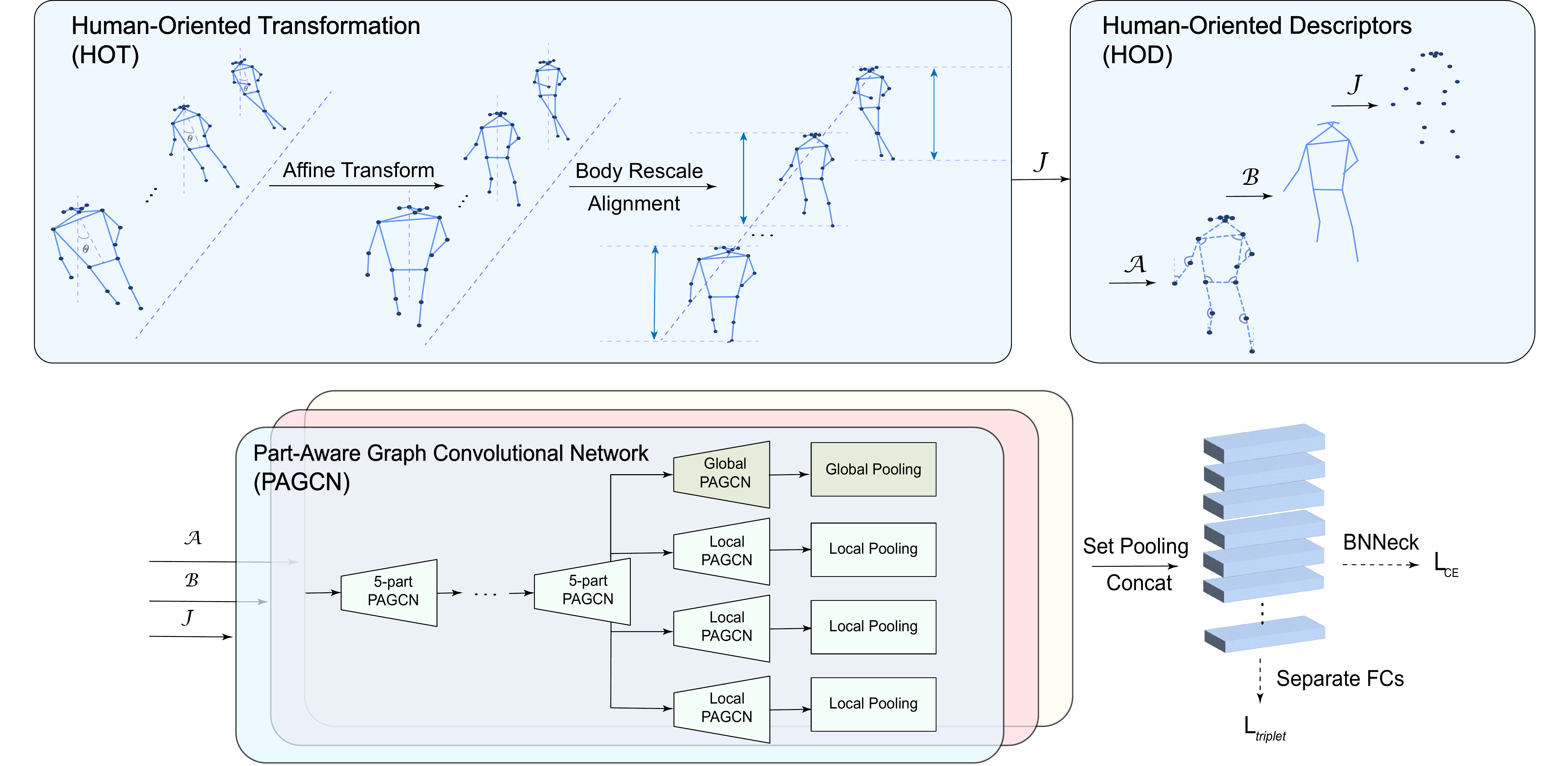}
    \caption{The framework of GPGait. The original pose sequence is first transformed into a unified representation by Human-Oriented Transformation. Then, angle, bone and joint features generated by Human-Oriented Descriptors are learned dependently through a multi-branch network named PAGCN. The output features are concatenated in part dimension and learned by a fully connected layer separately. Finally, a widely used BNNeck~\cite{luo2019strong} is adopted to adjust feature space. Triplet loss~\cite{hermans2017defense} and cross-entropy loss are utilized to supervise the whole training process. }
    \label{fig:pipeline}
\end{figure*}

\setParDis
{\setlength{\parindent}{0cm}
    \textbf{Silhouette-based methods  }
    GaitSet~\cite{chao2019gaitset} aggregates temporal information in silhouette sequences using a statistical function to adapt to different frame rates. GaitPart~\cite{fan2020gaitpart} uses Focal Convolutional Layer to extract fine-grained local features and Micro-motion Template Builder with different window sizes to extract local temporal information. GaitGL~\cite{lin2021gait}  uses 3D Convolution to extract global and local spatiotemporal information.  LagrangeGait~\cite{chai2022lagrange} adds a local motion extractor and a viewpoint branch based on GaitGL to get more discriminative local temporal information.

    \setlength{\parindent}{0cm}
    \textbf{Pose-based methods  }
    PoseGait~\cite{liao2020model} employs 3D pose information to generate multi-feature vectors and uses a CNN to extract the gait information in both spatial and temporal dimensions. GaitGraph~\cite{teepe2021gaitgraph} and GaitGraph2~\cite{teepe2022towards} adopt Graph Convolutional Network for gait recognition, treating keypoints as nodes and limbs as edges to form a topology graph.
    GaitTR~\cite{zhang2022spatial} and GaitMixer~\cite{pinyoanuntapong2022gaitmixer} use the self-attention~\cite{vaswani2017attention} to explore long-range spatial correlations, and temporal convolution with a large kernel size to extract long temporal information.}
\setParDef
\subsection{Generalization in Gait Recognition}
The generalization ability of the model is a practically significant concern. From GaitSet~\cite{chao2019gaitset} to LagrangeGait~\cite{chai2022lagrange}, the size of all the silhouette data is aligned based on methods in~\cite{takemura2018multi}. Each silhouette is put in the central position to generate a uniform representation, which is more stable and robust than the original input. Cross-domain experiments  are done by~\cite{zheng2022gait} with the backbone of GaitSet between popular public datasets and Gait3D to prove the importance of in-the-wild datasets. GaitEdge~\cite{liang2022gaitedge} evaluates their method on cross-dataset settings to demonstrate the irrelevance to RGB texture and color. Self-supervised learning has also been employed by~\cite{fan2022learning} to improve the ability to generalize in unseen domains. In skeleton-based methods, PoseGait~\cite{liao2020model} and CNN-Pose~\cite{an2020performance} use the human neck to align the skeletons and spine length to rescale the body. However, the spine length is sensitive to occlusion which introduces noise and even errors. Rashmi \etal~\cite{rashmi2022human} uses the dataset-independent statistics to rescale the skeleton, which still lacks the ability to generalize for the scale changes of skeletons across datasets.

\section{Method}
\label{sec:method}
In this section, we first present the pipeline of GPGait. Then we display a Human-Oriented Transformation (HOT) and a series of Human-Oriented Descriptors (HOD)  which generate a unified and enriched pose representation, followed by the description of the Part-Aware Graph Convolutional Network (PAGCN) that is specially designed for efficient fine-grained feature extraction.
\subsection{Pipeline}
As shown in Fig.\ref{fig:pipeline}, the pose sequences are first fed into Human-Oriented Transformation (HOT) to get a unified pose representation. Then a series of Human-Oriented Descriptors (HOD) are obtained to generate discriminative and domain-invariant features.
Due to the different distributions of joints, angles and bones,
we perform different feature extractions with a parameter-independent multi-branch architecture. The multi-features are learned through  Part-Aware Graph Convolutional Network (PAGCN) which can enable efficient human graph partitioning and local-global relationship building.  The pooling operations on semantic body parts are utilized at the end of the network to get the final embedding for recognition. We use the separate triplet and cross-entropy  losses to supervise the training process.
\subsection{Human-Oriented Transformation}
In this part, as Fig.\ref{fig:pipeline} shows, we transform the original data from various cameras to a stable and unified pose representation. Specifically, HOT consists of three phases of affine transform, body rescale and alignment to eliminate the environment covariance like viewpoints, distance away from  the camera, offset noises, \etal.

First, an affine transform  is used to overcome the problem of slant skeletons resulting from different camera views. Formally,
we suppose the original 2D frame set of one subject sequence is
\begin{math}
    P = \{ p \in \mathbb{R}^{T_{in} \times V_{in} \times C_{in}}\}
\end{math}
, where \(T_{in}, V_{in}, C_{in}\) denote the number of input frames, joints, and coordinates, respectively.
We regard the neck \(\mathbf{p_{neck}}\) as the median position of the right shoulder and left shoulder. Similarly, the location of  hip joint  \(\mathbf{p_{hip}}\) is the average of the right hip  and the left hip.  The spine is considered as the line between the neck and hip.
We take the spine as the axis, and the neck as a center to transform the inclined skeleton perpendicular to the ground. The rotation angle $\theta$ is calculated as:
\begin{equation}\label{eq:angle}
    \theta = \arctan(\frac{p_{neck}^{c_x} - p_{hip}^{c_x}}{p_{neck}^{c_y} - p_{hip}^{c_y}}),
\end{equation}
where \(c_x\) is the coordinate of x-axis and \(c_y\) is the coordinate of y-axis.  Affine transform is applied  exclusively to sequences with serious slant problems when the angle  \(\theta\) is larger than the threshold \(\phi\). Otherwise, the original pose $P$ is directly adapted. We assume the concatenation of \(P\) and \(\mathbf{p_{neck}}\) in \(C\) dimension is  \(D=\{\mathbf{d_i}|i=1,2,...,V_{in} \}\).  The process can be formulated as:
\begin{equation}
    \label{eq:2}
    P_a =\left\{
    \begin{aligned}
         & M_{a}D, & \theta \geqslant \phi \\
         & P,      & {\rm otherwise }
    \end{aligned}
    \right.
\end{equation}
\begin{equation}
    M_{a} =
    \begin{bmatrix}
        \cos{\theta} & -\sin{\theta} & 1-\cos{\theta} & \sin{\theta}   \\
        \sin{\theta} & \cos{\theta}  & -\sin{\theta}  & 1-\cos{\theta}
    \end{bmatrix},
\end{equation}
\begin{equation}
    \mathbf{d_i}=
    \begin{bmatrix}
        p_{i}^{c_x} & p_{i}^{c_y}    & p_{neck}^{c_x}
                    & p_{neck}^{c_y}
    \end{bmatrix}^\top,
\end{equation}
where \(P_{a}\) is the output of the affine transform, \(M_{a}\) is the affine transform matrix, \(p_{i}^{k}\) denotes the \(i\)-th joint at the \(k\)-th coordinate, and \(k \in \{c_{x},c_{y}\}\).
After the series of operations, the slanted human spine in all the sequences is perpendicular to the ground.

Second, considering the scale of the skeleton in each frame is variational, we propose a simple yet effective body-rescale (in Eq.\ref{eq:center1}) method to achieve a uniform height of  skeletons by dividing the difference between the maximum and minimum keypoint values along the vertical axis.
\begin{equation}
    \label{eq:center1}
    \mathbf{p_i^{\prime}} =  \mathbf{pa_i}*\frac{h_{unif}}{\max(\mathbf{pa^{c_y}})-\min(\mathbf{pa^{c_y}})},
\end{equation}
where \(\mathbf{pa}\) is taken from \(P_a\) and \(h_{unif}\) is a factor to control the uniform body height.

Third, an alignment operation (in Eq.\ref{eq:center2}) is performed on the human joints, in which the camera coordinate is converted to a human-oriented coordinate:
\begin{equation}
    \label{eq:center2}
    \mathbf{j_{i}} = \mathbf{p_{i}^{\prime}} - \mathbf{p_{neck}^{\prime}}.
\end{equation}
The  unified pose sequence we get in this module is \(\mathcal{J}=\{\mathbf{j_{i}}|i=1,2,...,V_{in} \}\) .

\begin{figure}[]
    \centering
    \includegraphics[width=\linewidth]{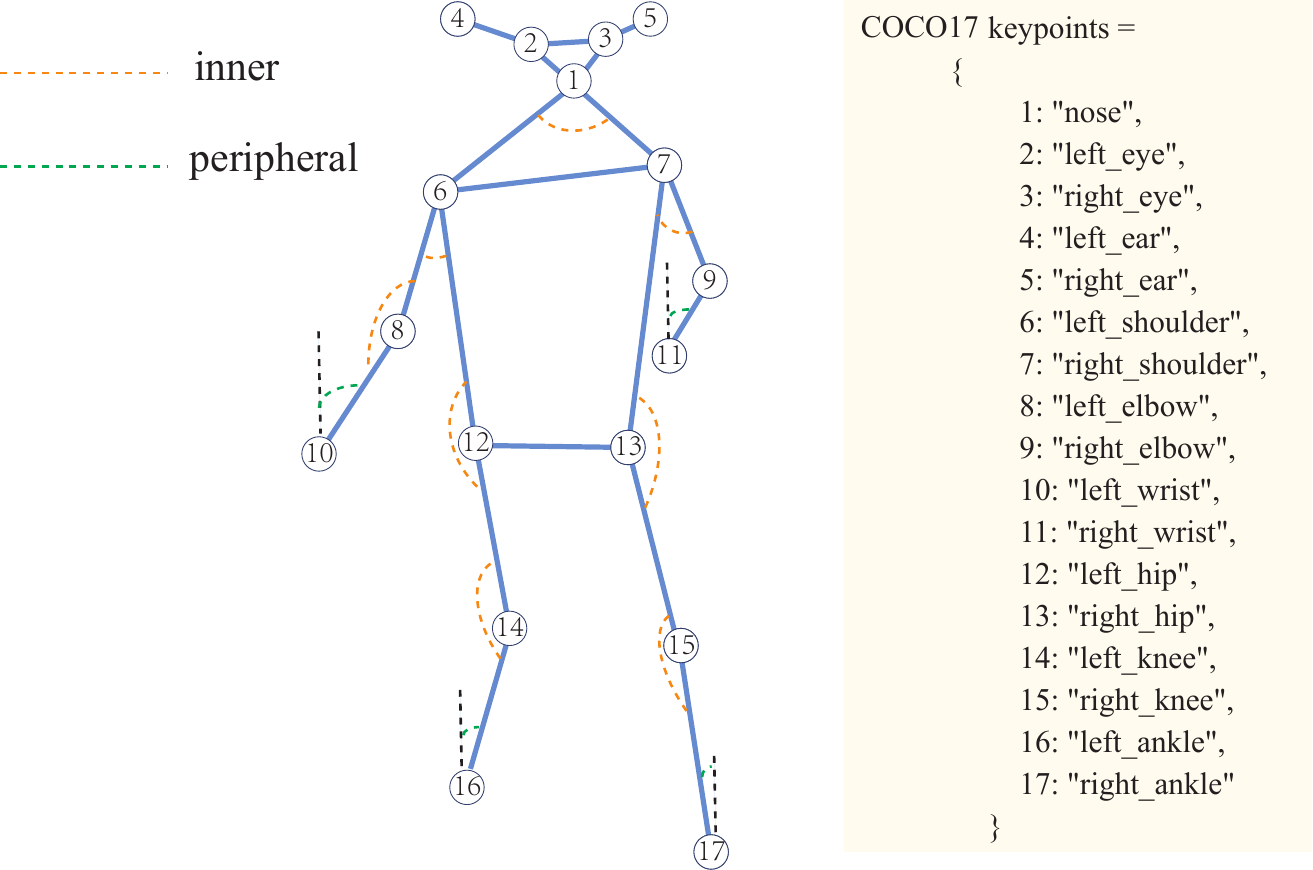}
    \caption{The detailed description of COCO2017 format with 17 keypoints. The figure also illustrates the sketch of inner and peripheral angles calculated in Human-Oriented Descriptors.}
    \label{fig:COCO17}
\end{figure}

\subsection{Human-Oriented Descriptors}
Recent works~\cite{liao2020model,teepe2022towards} have noticed that multi-input characterization can lead to improved recognition performance.
However, these works tend to underestimate the potential of the second-order information related to bones and angles. Specifically, the input of bone and angle features usually have different distributions from that of joint features, which can result in a totally different relation-learning process in GCN. The direct input of joints or the earlier fusion leads to distinct degradation in performance.
So we explicitly add information of bone and angle to learn discriminative gait signatures.

For the  bone features (in Eq.\ref{eq:bone}), we consider the human bone as vectors, i.e., \(\mathcal{B}=\{\mathbf{b_i}| i=1,2,...,V_{in}\}\).
\begin{equation}\label{eq:bone}
    \mathbf{b_{i}} = \mathbf{j_{i}} - \mathbf{j_{adj(i)}},
\end{equation}
where \(adj(i)\) denotes the adjacency joint of \(i\)-th joint.
For angles of the skeletons, unlike previous works~\cite{song2020stronger,teepe2022towards} using angles between bones and the horizontal or vertical line, we design a human-oriented angle-calculating method. Specifically, we use inner angles and peripheral angles shown in Fig.\ref{fig:COCO17}, i.e., \(\mathcal{A}=\{a_i | i=1,2,...,V_{in}\}\).
\begin{equation}
    \label{eq:angle}
    a_i=\left\{
    \begin{aligned}
         & \arccos(\frac{s_{l}^{2}(i)+s_{r}^{2}(i)-s_{opp}^{2}(i)}{2*s_{l}(i)*s_{r}(i)}), \quad\mathrm{inner}   & \\
         & \arctan(\frac{j_{i}^{c_x}-j_{adj(i)}^{c_x}}{j_{i}^{c_y}-j_{adj(i)}^{c_y}}), \quad\mathrm{peripheral} &
    \end{aligned}
    \right.
\end{equation}
where the \(s_{l}(i),s_{r}(i),s_{opp}(i)\) denotes the length of adjacent sides and the opposite side of the  \(i\)-th joint. \(adj(i)\) denotes the adjacency joints of peripheral skeleton joints and \(j_{i}^{k}\) denotes  \(i\)-th joint at  \(k\)-th coordinate.
Specifically, taking the $\textcircled{\scriptsize{14}}$ joint in Fig.\ref{fig:COCO17} as an example, $[s_l, s_r]$ denotes the length of $[\textcircled{\scriptsize{14}}$-$\textcircled{\scriptsize{12}}, \textcircled{\scriptsize{14}}$-$\textcircled{\scriptsize{16}}]$,
while $s_{opp}$ denotes the length of $\textcircled{\scriptsize{12}}$-$\textcircled{\scriptsize{16}}$.

\subsection{Part-Aware Graph Convolutional Network}
\paragraph{PAGCN Block}

\begin{figure}
    \centering
    \includegraphics[width=\linewidth]{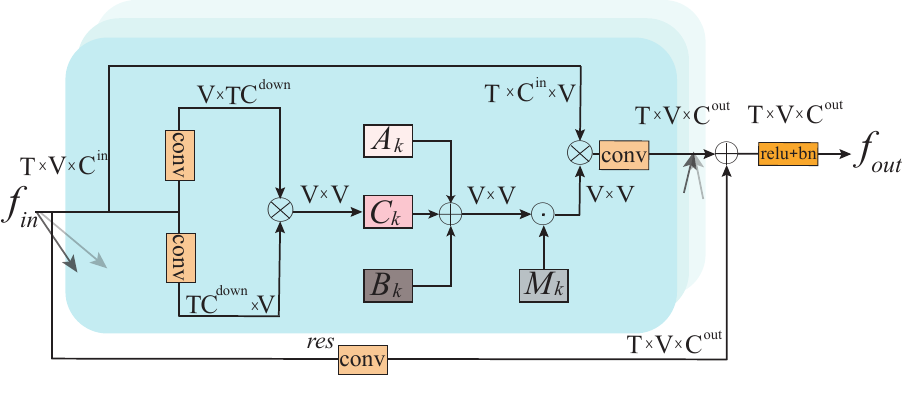}
    \caption{The detailed structure of Part-Aware Graph Convolutional Network (PAGCN) block.}
    \label{fig:MPAGCN}
    \vspace{-0.2cm}
\end{figure}
\begin{figure}[tp]
    \centering
    \includegraphics[width=\linewidth]{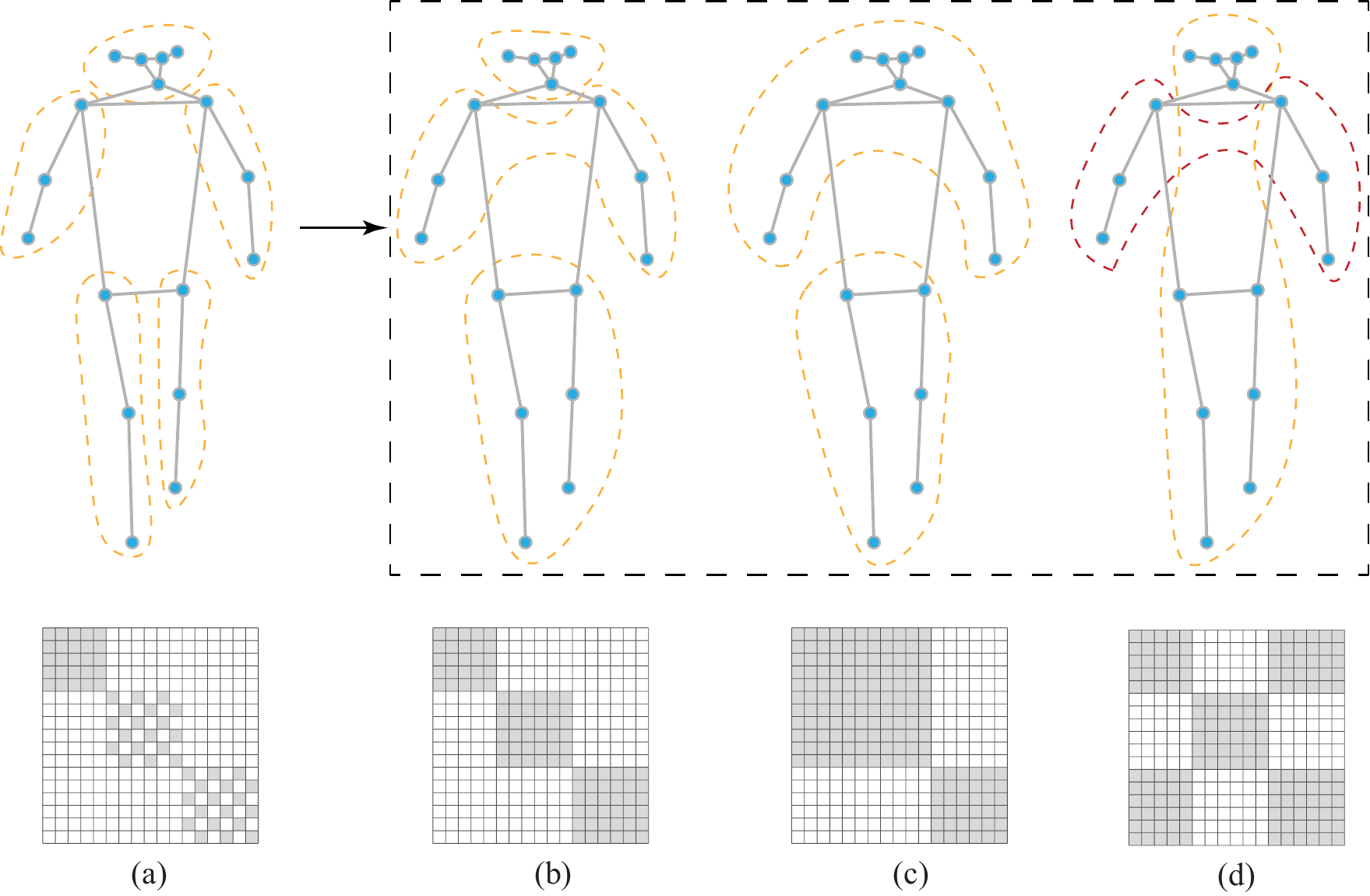}
    \caption{The partition masks following the human body structure. The larger body parts in (b), (c) and (d) are the combinations of  5 small parts in (a). The small black boxes in the adjacency matrices of keypoints refer to 1 and the white ones refer to 0.}
    \label{fig:mask}
    \vspace{-0.2cm}
\end{figure}

Although we get a unified representation to eliminate the camera covariance across datasets after HOT and HOD, there are actually not so many changes in body scale and offset when the subject is moving. So, it is important to capture the local variation and the overall structural information for the network. However, the traditional methods using graph convolutional network~\cite{teepe2021gaitgraph,teepe2022towards} or spatial transformer~\cite{zhang2022spatial} only construct the global relations between the keypoints, in which the slight move of some keypoints around the aligned center would be ignored. According to this, we propose a Part-Aware Graph Convolutional Network (PAGCN) to get local-global representations by using different partitioned masks. Our PAGCN block can be formulated in Eq.\ref{eq:magcn} and described in Fig.\ref{fig:MPAGCN}. The partitioned masks following the structure of the human body are shown in Fig.\ref{fig:mask}.
\begin{equation}
    \label{eq:magcn}
    f_{out} = \sum_{k}^{K_v} W_{k} (f_{in} (A_{k}+B_{k}+C_{k}) \odot M_{k}),
\end{equation}
where \(A_{k}\) denotes the predefined adjacency matrix of natural human structure, \(B_{k}\) denotes the parameterized adjacency matrix which can be updated in an end-to-end learning manner, \(C_k\) denotes the self-attention adjacency matrix which is used to construct global connections of joints for each sequence, the \(K_v\) denotes the number of graph subset. \(\odot\) stands for element-wise product. Especially, the value of \(M_k\) depends on the different body partition strategies \(G\), as shown in:
\begin{equation}
    \label{eq:m}
    M_{k}(i_1,i_2) =\left\{
    \begin{aligned}
         & 1, & p_{i_{1}},p_{i_{2}}\in g, g \subset G \\
         & 0, & \mathrm{otherwise}
    \end{aligned}
    \right.
\end{equation}
where \(G=\{g_1,g_2,..,g_n\}\), \(n\) means the number of different body partition strategies shown in Fig.\ref{fig:mask}. And the global PAGCN block is with $M_{k}(i_1,i_2)=1$.
Different from the previous work~\cite{shi2019two}, PAGCN allows for explicit graph partition and unleashes the deep GCN's expressive power in the training phase through mask operations, which is verified in the experimental study.

\paragraph{PAGCN Backbone}
The design of the backbone is mainly based on the following considerations: \textbf{1)} One part cannot include too few keypoints because each of them contains limited information. \textbf{2)} In order to make each part as distinguishable as possible, we try to avoid including the information of keypoints outside of the part during the process of learning the relationships . \textbf{3)} The backbone should be light and efficient. \textbf{4)} To adapt to diverse human feature distributions and enhance performance, multiple descriptor features should be applied using different parameters. Motivated by these principles, the overall backbone is composed of three  parameter-independent branches. And in each branch, we stack 5-part PAGCN blocks with shared parameters in the first few layers to construct relations between keypoints in small parts as shown in Fig.\ref{fig:pipeline}.
In the last few layers, we use larger-part PAGCN blocks to learn the relationships inside the major body parts and ensure better learning of part-specific relationships by avoiding parameter sharing across different partition masks. The whole architecture of the network can not only learn the fine-grained body part features but also prioritizes significant consideration of the second-order joint information.

In previous works, using split feature maps is considered effective for silhouette-based methods in gait recognition tasks. Instead of pyramid mapping~\cite{chao2019gaitset,fu2019horizontal}, we use semantic body feature mapping for the final recognition since every keypoint has its own semantic information and each body part has specific abilities for identification. We perform the pooling operation on the predefined body parts and on the whole body features to get the final embedding. We assume that \( f_{M} \in \mathbb R^{T \times V \times C}\) is the output of a last-layer PAGCN block, \(C\) is the output channels, \(T\) is the number of frames in a sequence and \(V\) is the number of keypoints. The feature mapping in spatial dimension can be formulated as:
\begin{equation}
    \label{eq:v_pooling}
    f_{vp}^{n} = {\rm ap}_{n}(f_{M})+{\rm mp}_{n}(f_{M}),
\end{equation}
where \( {\rm ap}(\cdot) \) refers to the statistical function \( average\) \(pooling \) and \( {\rm mp}(\cdot) \) refers to \( max\) \(pooling \) along the \(V\) dimension. \( n \) is the \( n \)-th body part. The output \(f_{vp}^{n} \in \mathbb R^{T \times 1 \times C} \). Then, a set pooling function is used to aggregate the temporal information:
\begin{equation}
    \label{eq:t_pooling}
    f_{tp}^{n} = {\rm sp}(f_{vp}^{n}),
\end{equation}
where \( {\rm sp}(\cdot) \) is the \( max\) \(pooling \) function along the \(T\) dimension and \(f_{tp}^{n} \in \mathbb R^{1 \times 1 \times C} \). The final representation for recognition is to concatenate the output of each part from three branches including \emph{joint}, \emph{angle} and \emph{bone} branch:
\begin{equation}
    \label{eq:t_pooling}
    F_{branch} = {\rm cat}(f_{tp}^{1},...,f_{tp}^{n}),
\end{equation}
\begin{equation}
    \label{eq:final}
    F_{final} = {\rm cat}(F_{joint},F_{angle},F_{bone}),
\end{equation}

\subsection{Optimization}
During training, the triplet loss~\cite{hermans2017defense} \(L_{triplet}\) and the cross-entropy loss \(L_{CE}\) are calculated for each part independently and the average of all parts  is used to supervise the learning process of the model:
\begin{equation}
    \label{eq:loss}
    L_{joint} = L_{triplet} + \gamma L_{CE},
\end{equation}
where the hyper-parameter \(\gamma\) is used to balance the weights of two losses.

\begin{table*}[]
    \begin{center}

        \caption{Rank-1 accuracy (\%) on four popular datasets: cross-domain and single-domain performance of GPGait and recent state-of-the-art pose-based methods, excluding identical-view case of indoor datasets. \protect\footnotemark[1]}
        \vspace{0.15cm}
        \fontsize{6}{8}\selectfont
        \resizebox{160mm}{!}{
            \begin{tabular}{c|c|ccccccc}
                \hline
                \multirow{3}{*}{\begin{tabular}[c]{@{}c@{}}Source \\      Dataset\end{tabular}} & \multirow{3}{*}{Method}            & \multicolumn{7}{c}{Target Dataset}                                                                                                                                                                                                                                     \\ \cline{3-9}
                                                                &                                    & \multicolumn{4}{c|}{CASIA-B}       & \multicolumn{1}{c|}{\multirow{2}{*}{OUMVLP-Pose}} & \multicolumn{1}{c|}{\multirow{2}{*}{GREW}} & \multirow{2}{*}{Gait3D}                                                                                                          \\ \cline{3-6}
                                                                &                                    & NM                                 & BG                                                & CL                                         & \multicolumn{1}{c|}{Mean}           & \multicolumn{1}{c|}{}               & \multicolumn{1}{c|}{}               &                \\ \hline
                \multirow{4}{*}{CASIA-B}                        & GaitGraph\cite{teepe2021gaitgraph} & 86.37                              & 76.5                                              & 65.24                                      & \multicolumn{1}{c|}{76.04}          & \multicolumn{1}{c|}{0.07}           & \multicolumn{1}{c|}{0.45}           & 0.90           \\
                                                                & GaitGraph2\cite{teepe2022towards}  & 80.29                              & 71.40                                             & 63.80                                      & \multicolumn{1}{c|}{71.83}          & \multicolumn{1}{c|}{0.07}           & \multicolumn{1}{c|}{0.48}           & 1.10           \\
                                                                & GaitTR\cite{zhang2022spatial}      & \textbf{94.72}                     & \textbf{89.29}                                    & \textbf{86.65}                             & \multicolumn{1}{c|}{\textbf{90.22}} & \multicolumn{1}{c|}{0.07}           & \multicolumn{1}{c|}{0.62}           & 1.10           \\
                                                                & GPGait(ours)                       & 93.60                              & 80.15                                             & 69.29                                      & \multicolumn{1}{c|}{81.01}          & \multicolumn{1}{c|}{\textbf{2.84}}  & \multicolumn{1}{c|}{\textbf{9.97}}  & \textbf{8.90}  \\ \hline
                \multirow{4}{*}{OUMVLP-Pose}                    & GaitGraph\cite{teepe2021gaitgraph} & 4.85                               & 4.84                                              & 3.90                                       & \multicolumn{1}{c|}{4.53}           & \multicolumn{1}{c|}{4.24}           & \multicolumn{1}{c|}{0.67}           & 1.50           \\
                                                                & GaitGraph2\cite{teepe2022towards}  & 8.83                               & 7.62                                              & 5.13                                       & \multicolumn{1}{c|}{7.19}           & \multicolumn{1}{c|}{\textbf{70.68}} & \multicolumn{1}{c|}{0.85}           & 1.40           \\
                                                                & GaitTR\cite{zhang2022spatial}      & 10.10                              & 8.26                                              & 5.17                                       & \multicolumn{1}{c|}{7.84}           & \multicolumn{1}{c|}{39.77}          & \multicolumn{1}{c|}{1.06}           & 2.60           \\
                                                                & GPGait(ours)                       & \textbf{44.36}                     & \textbf{31.97}                                    & \textbf{22.35}                             & \multicolumn{1}{c|}{\textbf{32.90}} & \multicolumn{1}{c|}{59.11}          & \multicolumn{1}{c|}{\textbf{11.13}} & \textbf{9.00}  \\ \hline
                \multirow{4}{*}{GREW}                           & GaitGraph\cite{teepe2021gaitgraph} & 10.54                              & 7.73                                              & 5.73                                       & \multicolumn{1}{c|}{8.00}           & \multicolumn{1}{c|}{0.17}           & \multicolumn{1}{c|}{10.18}          & 4.40           \\
                                                                & GaitGraph2\cite{teepe2022towards}  & 8.85                               & 7.18                                              & 5.13                                       & \multicolumn{1}{c|}{7.05}           & \multicolumn{1}{c|}{0.22}           & \multicolumn{1}{c|}{34.78}          & 8.30           \\
                                                                & GaitTR\cite{zhang2022spatial}      & 7.60                               & 6.36                                              & 6.40                                       & \multicolumn{1}{c|}{6.79}           & \multicolumn{1}{c|}{0.06}           & \multicolumn{1}{c|}{48.58}          & 7.30           \\
                                                                & GPGait(ours)                       & \textbf{57.87}                     & \textbf{45.98}                                    & \textbf{24.23}                             & \multicolumn{1}{c|}{\textbf{42.69}} & \multicolumn{1}{c|}{\textbf{4.25}}  & \multicolumn{1}{c|}{\textbf{57.04}} & \textbf{18.50} \\ \hline
                \multirow{4}{*}{Gait3D}                         & GaitGraph\cite{teepe2021gaitgraph} & 16.47                              & 12.18                                             & 8.29                                       & \multicolumn{1}{c|}{12.31}          & \multicolumn{1}{c|}{0.27}           & \multicolumn{1}{c|}{3.14}           & 8.60           \\
                                                                & GaitGraph2\cite{teepe2022towards}  & 12.32                              & 9.93                                              & 5.43                                       & \multicolumn{1}{c|}{9.23}           & \multicolumn{1}{c|}{0.09}           & \multicolumn{1}{c|}{2.39}           & 11.20          \\
                                                                & GaitTR\cite{zhang2022spatial}      & 4.50                               & 3.90                                              & 3.96                                       & \multicolumn{1}{c|}{4.12}           & \multicolumn{1}{c|}{0.06}           & \multicolumn{1}{c|}{4.38}           & 7.20           \\
                                                                & GPGait(ours)                       & \textbf{48.83}                     & \textbf{40.26}                                    & \textbf{19.43}                             & \multicolumn{1}{c|}{\textbf{36.17}} & \multicolumn{1}{c|}{\textbf{2.79}}  & \multicolumn{1}{c|}{\textbf{11.02}} & \textbf{22.40} \\ \hline
            \end{tabular}
        }
        \label{table:main}
        \vspace{-0.4cm}
    \end{center}
\end{table*}

\section{Experiments}
\label{sec:experiments}

{
    \subsection{Datasets and Implementation Details}
    To validate the effectiveness of GPGait, we conduct experiments under indoor (CASIA-B and OUMVLP-Pose) and  outdoor (Gait3D and GREW) scenarios, respectively.
    \setParDis

    \setlength{\parindent}{0cm}
    \textbf{CASIA-B~\cite{yu2006framework}} contains 124 subjects and each subject is required to walk under three conditions. For each condition, 11 sequences from multiple viewpoints range from 0\degree \ to 180\degree.
    During the testing stage, sequences of four normal conditions
    are regarded as the gallery, and the rest sequences are regarded as the probe. HRNet~\cite{8953615} is used to extract pose data from the RGB videos following the CASIA-B release  agreement.

    \setlength{\parindent}{0cm}
    \textbf{OUMVLP-Pose~\cite{an2020performance}} is based on the large gait recognition dataset OUMVLP~\cite{takemura2018multi} which contains 10,307 subjects, and the sequences of each subject are collected from 14 views (0 \degree, 15\degree, ..., 90\degree; 180\degree, 195\degree, ..., 270\degree). There are 2 sequences (\#00-01) under each view. 5,153 subjects are split for training and 5,154 subjects are split for testing. During the test, sequences with index \#01 are regarded as the gallery and those with index \#00 are used as the probe.

    \setlength{\parindent}{0cm}
    \textbf{Gait3D~\cite{zheng2022gait}} is a wild dataset containing 4,000 subjects with over 25,000 sequences. The data is collected in a supermarket from 39 cameras. 3,000 subjects are used for training and the rest for testing. At the testing stage, one sequence of each subject is selected to build the probe, and the rest sequences become the gallery.

    \setlength{\parindent}{0cm}
    \textbf{GREW~\cite{zhu2021gait}} is also a wild dataset that includes almost 3,500 hours   from 882 cameras. GREW dataset is split into a training set, a validation set and a testing set which contains 20,000, 345 and 6,000 subjects respectively.
    During testing, two sequences of each subject are regarded as the probe and the other two sequences are regarded as the gallery.

    \setlength{\parindent}{0cm}
    \textbf{Implementation Details  }
    In all experiments, \(h_{unif}\) is set to a fixed number of 225. The threshold \(\phi\) in Eq.\ref{eq:2} is set to 0.1 radians.
    We design the network capacity referring to the baselines \cite{teepe2021gaitgraph,teepe2022towards,zhang2022spatial} and does not deliberately tune it.
    Specifically, for CASIA-B and Gait3D, the number of 5-part blocks is 3 and the vector sizes of each block are (64, 64, 128).
    While for OUMVLP and GREW, the number of 5-part blocks is 4 and the vector sizes of each block are (64, 128, 128, 128, 128).
    During training, the length of pose sequences is fixed to 30 for OUMVLP and 60 for others in an unordered selecting manner.
    For data augmentations, we apply left-right flipping of the skeleton with a probability of 0.01 and gaussian noises are added to each keypoint with a probability of 0.3. The optimizer is adam~\cite{kingma2014adam} with a one-cycle learning rate schedule~\cite{smith2019super} of three phases, where initial, maximum, and final learning rates are set to 1e-5, 1e-3, and 1e-8. We adjust the batch size and the number of iterations to fit different dataset scales. \textbf{1)} In CASIA-B, we train the model for 40k iterations with a batch size of (4, 32). \textbf{2)} In Gait3D, we train the model for 60k iterations with a batch size of (32, 4).  \textbf{3)} In OUMVLP and GREW, the model is trained for 150k iterations with a batch size of (32, 16), (32, 8), respectively.
    During the test stage, all the frames of a sequence are fed into the network.
    \setParDef
}

\subsection{Performance Comparison}
{
    {\setlength{\parindent}{0cm}
            \textbf{Evaluation for Cross-Domain Settings\ }
            For a comprehensive comparison, we conduct cross-domain experiments on previous pose-based methods to evaluate the ability of generalization, i.e., train the model on the source dataset (Source) and test on other datasets (Target).
            As shown in Tab.\ref{table:main}, our method outperforms the existing pose-based methods over all of the source-target dataset pairs, suggesting our model's excellent generalization ability.
            In particular, compared to the second-best methods trained on the outdoor datasets, we achieve state-of-the-art results with considerable margins of 23.86\%, 34.69\% Rank-1 on Gait3D$\xrightarrow{}$CASIA-B and GREW$\xrightarrow{}$CASIA-B respectively.
            We have also surpassed the second-best result trained on indoor datasets by a significant margin of 9.35\%, 10.07\% Rank-1 on CASIA-B$\xrightarrow{}$GREW and OUMVLP-Pose$\xrightarrow{}$GREW respectively. }

    Furthermore, the cross-domain results of GPGait are found to even outperform or approach the source-domain results of recent methods. For instance, the cross-domain result (GREW$\xrightarrow{}$Gait3D) of GPGait outperforms the best source-domain results (Gait3D$\xrightarrow{}$Gait3D) of other methods with a margin of 7.3\% in Rank-1 accuracy. These superior performances collectively  demonstrate  the generalization capabilities  of our proposed method.

    \setParDis
    \setlength{\parindent}{0cm}
    \textbf{Comparison on Source Domain\ }
    The comparison in Tab.\ref{table:main} indicates that GPGait achieves nearly comparable or even better performance than the recent pose-based method in most cases. Especially for Gait3D, GPGait outperforms previous works by a considerable margin of 11.2\%.  Also, compared with other methods that failed to produce a stable result on specific datasets, GPGait achieves a relatively stable performance without any distinct performance degradation on single-domain settings across all four datasets.
    A detailed analysis of the performance on the source domain is included in Sec.\ref{sec:dis}.
    \setParDef
}

\footnotetext[1]{
    (a) We take great efforts to build a unified framework for pose-based gait recognition named FastPoseGait~(\url{https://github.com//BNU-IVC/FastPoseGait}) and re-run these experiments for a more fair comparison by sticking to the original implementations as much as possible.
    The results are a little different but comparable to those in the initial submission as well as those in the corresponding papers. \\
    \indent \indent (b) In the literature, there are two versions of pose-based CASIA-B estimated by HRNet~\cite{8953615} and SimCC~\cite{li2022simcc} respectively.
    Given that Gait3D and GREW are generated by HRNet, we finally use the HRNet version of CASIA-B for a unified experimental setting.
    We will also provide some results for the SimCC version of CASIA-B in our codebase. \\
    \indent \indent (c) For OUMVLP-Pose, the sequences are generated by AlphaPose~\cite{alphapose} consisting of 18 keypoints for each frame.
    In our experiments, we transform the keypoints into the COCO2017 format with 17 keypoints for the cross-domain evaluation.
}

\subsection{Ablation Study}
{
    To verify the effectiveness of the components in GPGait, a series of ablation studies are conducted on CASIA-B and Gait3D to show the source-domain and cross-domain performance of indoor and outdoor evaluation systematically.

    \setParDis
    {\setlength{\parindent}{0cm}
        \textbf{Analysis  of Human-Oriented Transformation }
        In this section, we first analyze the  effectiveness of HOT compared with other normalization methods. Then we insert HOT into different backbones to verify  generalization ability after normalizing the pose data. }

    \setParDef
    As shown in Tab.\ref{table:HCT}, compared with previous methods, the results of GPGait demonstrate the best generalization ability. \textbf{a) } Compared with spine-unit normalization~\cite{liao2020model,an2020performance}, HOT outperforms it by a large margin in the source domain as well. On the one hand, it proves the effectiveness of employing such a simple method to achieve remarkable results. On the other hand, the spine as a body part is  not a good solution for some special situations like occlusion which introduces abnormal calculations. \textbf{b)} Compared with the dataset-independent normalization method~\cite{rashmi2022human}, HOT demonstrates excellent generalization results. This is because the dataset statistics preserve the relative height of the human body in one dataset. However, the relative information is lost when the model is applied to another sequence recorded by a different camera of different datasets, which makes it less applicable in the real world.

    \begin{table}[]
        \caption{Analysis of different pose representations, in which we control the structure of the network.}
        \vspace{0.15cm}
        \fontsize{11}{15}\selectfont
        \resizebox{\linewidth}{!}{
            \begin{tabular}{c|cc|cc}
                \hline
                \multirow{2}{*}{Method}                           & \multicolumn{2}{c|}{CASIA-B$\rightarrow$Gait3D} & \multicolumn{2}{c}{Gait3D$\rightarrow$CASIA-B}                           \\ \cline{2-5}
                                                                  & Source                                          & Target                                         & Source & Target         \\ \hline
                HOT(ours)                                         & 81.01                                           & \textbf{8.90}                                  & 22.40  & \textbf{36.17} \\
                Spine-Unit~\cite{liao2020model,an2020performance} & 74.53                                           & 5.50                                           & 14.50  & 15.74          \\
                Dataset-Independent~\cite{rashmi2022human}        & 87.03                                           & 1.50                                           & 9.90   & 11.54          \\ \hline
            \end{tabular}
        }
        \vspace{-0.05cm}
        \label{table:HCT}
    \end{table}
    \begin{table}[]
        \caption{Analysis of HOT, in which we compare the performance of HOT on different backbones.}
        \vspace{0.15cm}
        \fontsize{6}{8}\selectfont
        \resizebox{\linewidth}{!}{
            \begin{tabular}{c|cc|cc}
                \hline
                \multirow{2}{*}{Method(w/wo)} & \multicolumn{2}{c|}{CASIA-B$\rightarrow$Gait3D} & \multicolumn{2}{c}{Gait3D$\rightarrow$CASIA-B}                           \\  \cline{2-5}
                                              & Source                                          & \multicolumn{1}{c|}{Target}                    & Source & Target         \\ \hline
                GaitGraph                     & 40.91                                           & 1.96                                           & 11.00  & 29.24          \\
                GaitGraph2                    & 45.66                                           & 3.20                                           & 12.20  & 24.49          \\
                GaitTR                        & 64.37                                           & 2.40                                           & 8.10   & 21.74          \\
                GPGait                        & 81.01                                           & \textbf{8.90}                                  & 22.40  & \textbf{36.17} \\ \hline
                GaitGraph                     & 76.04                                           & 0.90                                           & 8.60   & 12.31          \\
                GaitGraph2                    & 71.83                                           & 1.10                                           & 11.20  & 9.23           \\
                GaitTR                        & 90.22                                           & 1.10                                           & 7.20   & 4.12           \\
                GPGait                        & 86.15                                           & \textbf{2.70}                                  & 16.00  & \textbf{20.30} \\ \hline
            \end{tabular}
        }
        \label{table:HCT2}
        \vspace{-0.05cm}
    \end{table}
    In addition, to demonstrate the reusability of HOT on different backbones, we apply it to previous methods which are shown in Tab.\ref{table:HCT2}.  \textbf{a)} Although HOT reduces the performance of the previous pose-based methods on CASIA-B, its generalization ability remains relatively steady across different backbones. Also, all the methods with HOT on Gait3D improve by a large margin. This demonstrates that HOT can be effectively adapted to other works, enabling them to achieve better and more stable results.  \textbf{b)} For settings without HOT, GPGait still performs relatively stable and outperforms most previous works, which further demonstrates the superiority of PAGCN.

    \setParDis
    \setlength{\parindent}{0cm}

    \textbf{Analysis of Human-Oriented Descriptors }
    In Tab.\ref{tab:MFG}, the results show the impact of using different combinations of generated features in Human-Oriented Descriptors. It can be seen that using only a single modality as input can not yield satisfactory results. When combining the input types, corresponding improvements can be achieved in both the source domain and the target domain. Furthermore, the multi-features of joint, bone and angle can significantly boost the performance. The benefit of multi-features generated by HOD can explicitly describe discriminant information of the human body that includes human keypoints, human body structure, and gait movement.
    \begin{table}[]
        \caption{Analysis of  Human-Oriented Descriptors, in which we keep the network architecture consistent.}
        \vspace{0.15cm}
        \fontsize{11}{15}\selectfont
        \resizebox{\linewidth}{!}{
            \begin{tabular}{ccc|cc|cc}
                \hline
                \multicolumn{3}{c|}{Setting} & \multicolumn{2}{c|}{CASIA-B$\rightarrow$Gait3D} & \multicolumn{2}{c}{Gait3D$\rightarrow$CASIA-B}                                                                    \\ \hline
                Joint                        & Bone                                            & Angle                                          & Source         & Target        & Source         & Target         \\ \hline
                \textbf{ \checkmark}         & \textbf{}                                       & \textbf{}                                      & 77.10          & 7.20          & 17.90          & 31.12          \\
                \textbf{}                    & \textbf{ \checkmark}                            & \textbf{}                                      & 76.02          & 7.20          & 18.30          & 31.78          \\
                \textbf{}                    & \textbf{}                                       & \textbf{ \checkmark}                           & 43.78          & 2.90          & 5.30           & 14.18          \\
                \textbf{ \checkmark}         & \textbf{ \checkmark}                            & \textbf{}                                      & 80.08          & 7.70          & 19.20          & 32.46          \\
                \textbf{}                    & \textbf{ \checkmark}                            & \textbf{ \checkmark}                           & 79.40          & 7.70          & 18.30          & 32.74          \\
                \textbf{ \checkmark}         & \textbf{}                                       & \textbf{ \checkmark}                           & 77.42          & 7.00          & 17.00          & 34.46          \\
                \textbf{ \checkmark}         & \textbf{ \checkmark}                            & \textbf{ \checkmark}                           & \textbf{81.01} & \textbf{8.90} & \textbf{22.40} & \textbf{36.17} \\ \hline
            \end{tabular}
        }
        \label{tab:MFG}
        \vspace{-0.3cm}
    \end{table}

    \begin{table}[]
        \caption{Analysis of Multi-branch, in which we control the backbone of PAGCN. }
        \vspace{0.15cm}
        \fontsize{6}{8}\selectfont
        \resizebox{\linewidth}{!}{
            \begin{tabular}{c|cc|cc}
                \hline
                \multirow{2}{*}{Setting} & \multicolumn{2}{c|}{CASIA-B$\rightarrow$Gait3D} & \multicolumn{2}{c}{Gait3D$\rightarrow$CASIA-B}                                   \\ \cline{2-5}
                                         & Source                                          & Target                                         & Source         & Target         \\ \hline
                Single-Branch            & 75.54                                           & 7.10                                           & 18.90          & 33.04          \\
                Multi-Branch             & \textbf{81.01}                                  & \textbf{8.90}                                  & \textbf{22.40} & \textbf{36.17} \\ \hline
            \end{tabular}
        }
        \label{tab:Multi-branch}

    \end{table}

    \begin{table}[]
        \begin{center}
            \caption{Analysis of Partition, in which we control the whole network structure.}
            \vspace{-0.05cm}
            \fontsize{4.5}{6}\selectfont
            \resizebox{\linewidth}{!}{
                \begin{tabular}{c|cc|cc}
                    \hline
                    \multirow{2}{*}{Partition} & \multicolumn{2}{c|}{CASIA-B$\rightarrow$Gait3D} & \multicolumn{2}{c}{Gait3D$\rightarrow$CASIA-B}                                   \\ \cline{2-5}
                                               & Source                                          & Target                                         & Source         & Target         \\ \hline
                    w                          & \textbf{81.01}                                  & \textbf{8.90}                                  & \textbf{22.40} & \textbf{36.17} \\
                    w/o                        & 76.47                                           & 7.10                                           & 20.40          & 35.15          \\ \hline
                \end{tabular}
            }
            \label{tab:mask}
        \end{center}
        \vspace{-0.7cm}
    \end{table}

    \setlength{\parindent}{0cm}

    \textbf{Impact of Multi-Branch in PAGCN }
    This section aims to explore the contribution of multi-branch architecture to learn the features generated by HOD.
    In single-branch settings, we concatenate the three types of features in the channel dimension and put them into a one-branch network as the operations in GaitTR\cite{zhang2022spatial}. The multi-branch settings consist of parameter-independent branches to extract three types of features. In Tab.\ref{tab:Multi-branch}, the use of multi-branch networks can lead to better performance, showing its effectiveness in learning specific expressions. This is mainly because each branch can focus on extracting information from different types of features separately. In contrast, a single-branch network merges the information from different data distributions, making the learning process  challenging.

    \setlength{\parindent}{0cm}
    \textbf{Impact of Partition in PAGCN }
    A completely identical network without multiplying the \(M_k\) in Eq.\ref{eq:magcn} is designed to verify the effectiveness of partition masks, where the number of parameters  is completely the same as well. In Tab.\ref{tab:mask}, the remarkable performance of the  network leveraging masks proves that it is beneficial to extract the discriminative body features at both local and global levels with the help of partition on the adjacency matrices.
    \setParDef
}
\

\section{Discussion}
\label{sec:dis}
{
    \setlength{\parindent}{0cm}
    \textbf{Source-Domain Results   }
    It can be seen that GPGait does not achieve state-of-the-art but comparable source-domain results on some datasets. The main reason is that HOT eliminates certain relative information in one dataset. For example, the relative information of body height is a discriminative factor to improve single-domain performance. But it does not exist in real-world scenarios due to the various heights of camera viewpoints. HOT rescales all the skeletons to a uniform height, which can force the network more concerned about gait-relevant features and further enhance the practicality of skeleton-based methods. Therefore, it is reasonable for the performance degradation on some datasets, and we believe future research that incorporates advanced feature extraction methods under the human-oriented representation can improve the performance of pose-based methods on both source-domain and cross-domain settings.
    \setParDis

    \setlength{\parindent}{0cm}
    \textbf{Compared with Silhouette-based Methods  }
    Compared to recent silhouette-based methods, the performance of GPGait is still limited. An evident explanation is that the poses used in GPGait lose body shape compared with the silhouettes. But pose-based methods have their own strengths like the robustness to wearings and explicit modeling for proportions and relations of body parts, which deserves further and continuous exploration.

    \setlength{\parindent}{0cm}
    \textbf{Prospect }
    Pose as an important modality for human representation contributes a lot to gait recognition as well. However, the lack of ability to generalize limits the application and further development of skeleton-based gait recognition.
    Our GPGait framework takes both the input and method into account, proposing a viable solution to address the generalization problem of pose-based methods. We are expecting the unified representation of skeletons can be utilized in later works for a fair and applicable future of pose-based methods. And PAGCN offers a promising approach for establishing part relations and extracting fine-grained gait information, which can be readily adapted for future research.
    Overall, more advanced pose-based methods are expected to further narrow the gap between the lab and the real world and promote the development of gait recognition.
    \setParDef
}

\section{Conclusion}
\label{sec:conclusion}
In this paper, we present a generalized pose-based framework (GPGait), which transforms the arbitrary human pose into a unified representation and  make full use of human pose characteristics to extract multi-features in Human-Oriented Transformation and Human-Oriented Descriptors.
Part-Aware Graph Convolutional Network allows efficient partitions of the human graph and the effective learning of local-global relations. Experiments on four benchmarks (including indoor and outdoor scenarios) have indicated that GPGait achieves the highest accuracy on cross-domain settings and the most stable performance on single-domain settings, which also demonstrates the great potential of pose-based gait recognition.

\vspace{0.5em}

{\setlength{\parindent}{0 em}
    \textbf{Acknowledgement }
    This work is jointly supported by National Natural Science Foundation of China (62276025, 62206022), Fundamental Research Funds for the Central Universities (2021NTST31) and Shenzhen Technology Plan Program (KQTD20170331093217368).
}
\section{Supplementary Materials}
\subsection{Human-Oriented Descriptor of Angle}
Human-Oriented Descriptor of angle generates two types of skeleton angles to describe the gait movements, known as inner and peripheral. As shown in  Fig.\textcolor{red}{3}, inner angle is the angle between two adjacent bones of a joint, which describes angular changes of the skeleton inside human body. While peripheral ones are  formed between vertical lines and adjacent bones of joints on the outside, which reflect movements at the edge of skeletons. This calculation method does not rely on the absolute coordinate and is entirely human-oriented, which makes it more generalizable across different cameras and environments.

\subsection{Analysis of Over-Smooth}
\begin{figure}[h]
    \centering
    \includegraphics[width=\linewidth]{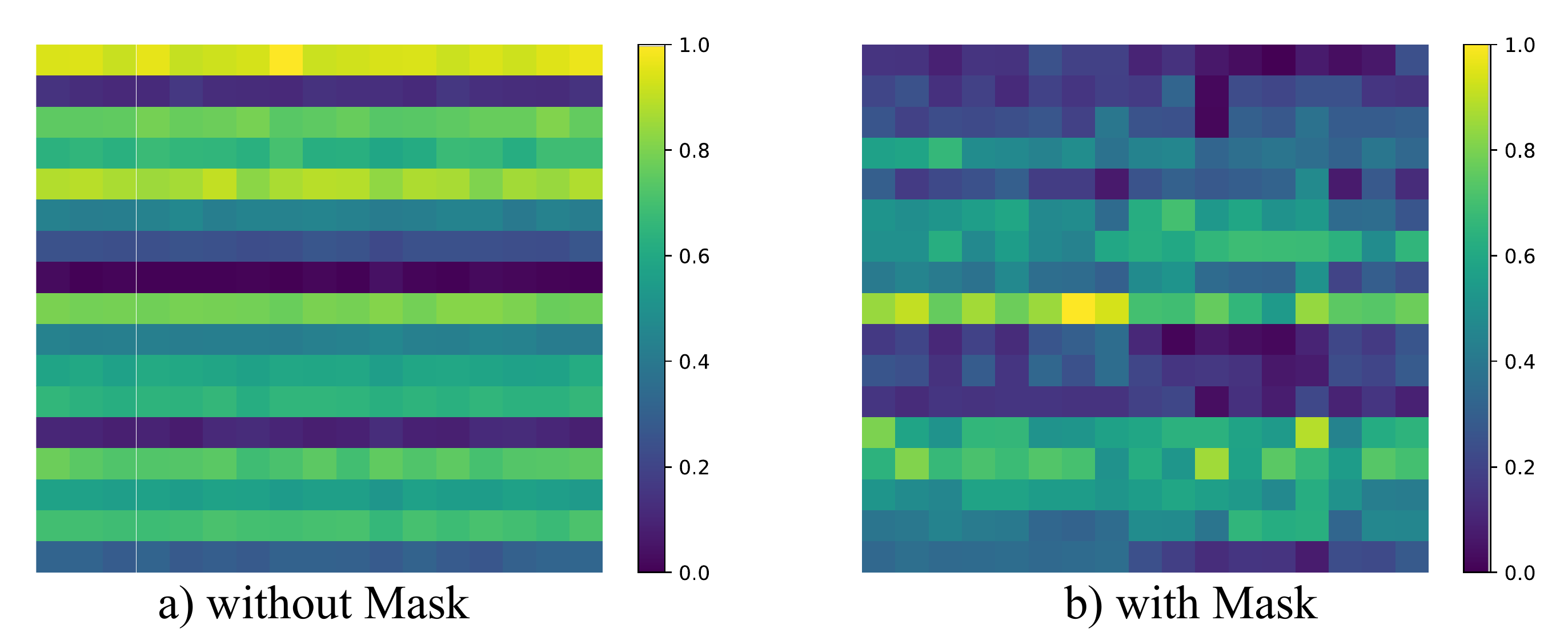}
    \caption{The visualization of the keypoint heatmaps. a) points to the network without Part-Aware mechanism, and b) points to the network with Part-Aware mechanism. The x-axis represents the keypoints (17) of human body, and the y-axis represents the channels (17) taken from the final embedding.}
    \label{fig:over_smooth}
\end{figure}

Over-smooth is a common problem in Graph Convolutional Networks~\cite{li2018deeper, chen2020simple,huang2020tackling}
, \ie, with the increase of non-linearity layers in GCN, the representation of each node in a connected component tends to converge to the same value. For the graph of human skeletons, each node contains its own semantic information and over-smooth is expected not to be so severe.
Our proposed Part-Aware method can restrict the interaction of keypoint information within parts, which helps alleviate over-smooth to a large extent.
By visualizing the heatmap of keypoints in the network, as shown in Fig.\ref{fig:over_smooth}, we can see that the value of each keypoint feature becomes more discriminative. This demonstrates the effectiveness of Part-Aware GCN blocks in reducing the over-smooth.


\vspace{-0.5em}
\subsection{More Discussion on Comparison}

\vspace{-0.05em}
First, as stated in Sec.\textcolor{red}{5}, GPGait does not achieve the highest performance (but comparable to state-of-the-art) on the source domain for some of the datasets.
The main reason is the adoption of Human-Oriented Transformation, aimed at achieving a unified  representation that can be compared across various cameras and scenes. During this transformation, certain attributes present exclusively in a single domain (\eg the relative height of the human body) are lost.
However, as far as we can see, a unified representation is essential to improve the generalization ability and achieve practical recognition, which is verified by a thorough cross-domain study.

\vspace{-0.05em}
Second, it can be observed from Tab.\textcolor{red}{1} that our method is the most stable one across different datasets. For example, GaitTR achieves remarkable results on CASIA-B, while its performance on OUMVLP-Pose and Gait3D is instead much inferior to GaitGraph2 (\eg GaitTR: 39.77\% \textit{v.s.} GaitGraph2: 70.68\% on OUMVLP-Pose). In comparison, GPGait achieves the highest or close to the highest accuracy on the source domain of each dataset, and more importantly, the cross-domain performance is improved by a large margin (\eg +34.69\% for GREW$\xrightarrow{}$ CASIA-B).

\vspace{-0.05em}
In summary, our work makes one of the pioneering attempts to improve the generalization ability of pose-based gait recognition, which is worthy of continuous attention due to its robustness to carrying and clothing. We have built a project to release GPGait as well as the re-implementations of improved baselines (GaitGraph~\cite{teepe2021gaitgraph}, GaitGraph2~\cite{teepe2022towards}, GaitTR~\cite{zhang2022spatial}), and we hope these efforts will promote pose-based research for gait recognition.

\subsection{Ethical Statements}
Our work is  guided by a strong commitment to upholding ethical and security standards\cite{iso} when handling biometric data, with the aim of promoting the development of gait recognition for the betterment of society and the improvement of human well-being.

{\small
\bibliographystyle{ieee_fullname}
\bibliography{egbib}
}

\end{document}